# Improving the portability of predicting students' performance models by using ontologies

Javier Lopez-Zambrano [1,3] · Juan A. Lara [2] · Cristóbal Romero [3]

1 Escuela Superior Politécnica Agropecuaria de Manabí (ESPAM MFL), Faculty of Computing,
SISCOM Group, 131106 Calceta, Ecuador
2 Department of Computer Science, Madrid Open University, UDIMA, Carretera de La Coruna,
KM. 38500, Vía de Servicio, No. 15, 28400 Collado Villalba, Madrid, Spain
3 Department of Computer Science and Numerical Analysis, University of Córdoba (UCO),
14071 Córdoba, Spain

**Abstract:** One of the main current challenges in Educational Data Mining (EDM) and Learning Analytics (LA) is the portability or transferability of predictive models obtained for a particular course so that they can be applied to other different courses. To handle this challenge, one of the foremost problems is the models' excessive dependence on the low-level attributes used to train them, which reduces the models' portability. To solve this issue, the use of high-level attributes with more semantic meaning, such as ontologies, may be very useful. Along this line, we propose the utilization of an ontology that uses a taxonomy of actions that summarises students' interactions with the Moodle learning management system. We compare the results of this proposed approach against our previous results when we used low-level raw attributes obtained directly from Moodle logs. The results indicate that the use of the proposed ontology improves the portability of the models in terms of predictive accuracy. The main contribution of this paper is to show that the ontological models obtained in one source course can be applied to other different target courses with similar usage levels without losing prediction accuracy.

**Keywords:** Educational Data Mining; Predictive Modelling; Student Performance; Transfer Learning; Model Portability; Ontology.

1. Introduction

In recent decades, one of the main educational milestones is the advent of a new form of learning called e-learning (electronic learning), based on the use of the internet and technology to support students' online education. Nowadays, this form of learning is becoming particularly important due to the limitations defined by the authorities to restrain the spread of pandemics such as the one caused by Covid-19. The use of e-learning poses important advantages including the enabling of a more flexible temporal and spatial interaction than other forms of learning. Besides, vast amounts of learning process data can be collected, since it is based on the use of Learning Management Systems (LMS). Moodle (Dougiamas and Taylor, 2008) is one of the most used LMS overall, because, among other advantages, it is free, open and there is an important community of users who support its development. Data recorded by Moodle, in particular those that reflect students' interactions with educational resources, can be of great interest and applicability for building student behavior models. To analyze these data, approaches such as Educational Data Mining (EDM) and Learning Analytics (LA) are useful (Romero et al., 2020). In EDM, a field whose purpose is the extraction of knowledge from educational data, there are well-defined problems that have been addressed by the scientific community, such as the prediction of students' performance (Romero and Ventura, 2008; Romero and Ventura, 2013). Recently, it is more frequent to find works that propose new approaches to analyzing educational data for a particular course. However, one of the due challenges is creating models for a particular course that can be useful when used in other courses (Baker, 2019). These are what we call transferable or portable models (Boyer and Veeramachaneni, 2015).

In our previous work (López-Zambrano et al., 2020), we obtained models generated from Moodle's logs data and we studied the degree of portability of the models between subjects, grouped by area of knowledge and by the usage level of platform resources. We used Moodle's native raw attributes which, in certain combinations of courses, led us to a certain loss in the portability of models since these low-level attributes are very dependent on each particular course. To overcome this limitation from our previous research, in this paper we present a new approach based on the use of resources from the semantic web area, in particular, ontologies (Tang and Fong, 2010; Fong et al., 2011). One of the most promising lines in this respect, particularly when analyzing logs of students' interactions with the LMS, is the categorization or taxonomy of attributes. In this regard, Bloom's taxonomy plays an important role. Bloom's taxonomy is a multi-tiered model of classifying thinking according to six cognitive levels of complexity which in this new version are: Remembering, Understanding, Applying, Analysing, Evaluating and Creating (Forehand, 2005). Based on this idea, some works have even defined correspondence between the levels of Bloom's taxonomy and the different actions conducted by students in Moodle (Rollins, 2010). Some authors (Cerezo et al., 2020) proposed a categorization of low-level attributes into different higher-level codifications, such as Executing, Planning, Learning, and Reviewing. Precisely, our research aims to evaluate the degree of portability of models built by using ontologies of interaction-with-the-platform attributes. To do so, we defined an ontology inspired by Bloom's taxonomy and based on the work by (Cerezo et al., 2020), with the purpose of conducting a comprehensive study to measure the degree of portability of the models built based on that ontology (denoted as ontological models), compared with a previous similar study conducted by the authors (López-Zambrano et al., 2020) in which we did not use ontologies but instead employed low-level Moodle attributes (denoted as non-ontological models). The models have been built from students' interactions with Moodle logs and the class attribute to predict is binary and represents whether or not the student will pass the course (Pass/Fail). In this work, the courses have been grouped according to the usage level of Moodle activities/resources. This approach has already been used in previous studies with satisfactory results (López-Zambrano et al., 2020). Taking all this into consideration, the global objective of this paper is to provide an answer for the research question below:

- Can the ontological models obtained in one (source) course be applied in other different (target) courses with a similar usage level without losing prediction accuracy?

The rest of the paper is organized as follows: Section 2 reviews the literature related to this research. Section 3 describes the data and the experiments. Section 4 includes and discusses the results obtained. Finally, Section 5 presents the conclusions and future lines of research.

## 2. Background

Achieving generalizable and portable models is still an important challenge in the area of EDM, in spite of the important advances made in the last few years (Boyer and Veeramachaneni, 2015; Gašević et al., 2016; Hunt et al., 2018; Ding et al., 2019). In fact, Baker (2019) has considered what he calls the "Generalizability" or "New York City and Marfa" problem as one of the main challenges for the future of EDM, which is explained in detail in (López-Zambrano et al., 2020).

To address this challenge, the resource of resources from the semantic web seems to be a promising line. The semantic web is an extension of the current web in which information is provided with a certain meaning, which makes cooperation and portability easier (Dhuria and Chawla, 2014). Fundamental resources from the semantic web are the ontologies, because they provide a common understanding of a domain. In particular, they may be interesting resources in the e-learning field (Al-Yahya et al., 2015).

In this regard, several particular works should be highlighted. In (Octaviani et al., 2015) they present a tool, called RDB2Onto, for creating ontologies from Moodle logs, but this work does not validate the utility of such an ontology. In (Castro and Alonso, 2012) they propose a general architecture for EDM in which there is an educational ontology, but they do not define or develop the ontology, only providing a general statement of it as a part of a higher-level architecture. There are even some works such as the one presented in (Chang et al., 2020) where they utilize data mining techniques (association in this case) to build ontology-driven tutoring models for intelligent tutoring systems (this is precisely the opposite process to ours since we use the ontology for a further data mining analysis).

These previous works present general approaches. Other more specific works bear greater similarity to our study because they define particular ontologies to facilitate the EDM process. We found some works where the ontology created is not focused on attributes of students' interaction with the LMS.

In (Marinho et al., 2010) they propose an ontology to model EDM tasks, techniques, and parameters. In (Grivokostopoulou et al., 2014) they propose an educational system that utilizes ontologies and semantic rules to enhance the quality of educational content (curriculum) and the learning activities delivered to each student. In (Nouira et al., 2019), they propose an ontological model for assessment analytics. And finally; in (Dorça et al., 2017), they present an approach for the automatic and dynamic analysis of learning object repositories in which ontology models the relationships between the attributes and learning styles of the learning objects.

Other related works are those that define ontologies to model data of students' interactions with LMS resources. In (El-Rady, 2020), they propose an ontology where the student is the main class from which a series of associations arise that are connected to other classes that model the students' data (education, profile, social activities, etc.). That ontology is used as a part of a validation process to predict student dropout rates. Other related works are based on the idea of organizing the interaction attributes as part of a kind of taxonomy. It is worth highlighting the work presented in (Cerezo et al., 2020), where they propose a process mining method for a self-regulated learning assessment, and make use of an ontology inspired by Bloom's taxonomy. In (Montenegro-Marin et al., 2011), they also propose an ontology based on the idea of taxonomy, but not restricted to interaction attributes, as they consider many other features, such as the curriculum design, productivity, management, and so on. However, they do not validate the utility of the ontology.

Considering all the previous works, and to the best of the authors' knowledge, our work presented in this paper is the first that analyses the power of ontologies as a resource that makes the portability of EDM models easier and, in particular, it is also the only one for that purpose which is based on data from the students' interactions with the LMS. Furthermore, it is the first research that depicts a comparative study against a previous non-ontological similar approach, with the purpose of demonstrating the performance improvement obtained when using ontologies. Both of these innovative aspects are the core contributions of this paper.

## 3. Materials and methods

In this section, we describe both the data used and the preprocessing tasks we applied to them in order to transform the raw data gathered from the Moodle logs to the high-level attributes of the proposed ontology. We also describe the experimentation that we carried out in order to address our research question.

### 3.1. Data and preprocessing

We have used the log data of 1840 Cordoba University students from 16 different courses taught by the Computer Science Department. Table 1 summarises these courses. For each course, it shows the subject or name of the course (Subject), our own identification Code, name of the Degree, Year in the degree/curriculum, number of students (#Users), and the level of Moodle Usage (Low, Medium or High). To accomplish the ethical and privacy issues about using these data, we have used informed consent with all the instructors and we have also anonymized all information about students (Pardo and Siemens, 2014).

**Table 1.** Information of all subjects.

| Subject | Code | Degree | Year | #Users | Moodle Usage |
|---|---|---|---|---|---|
| Introduction to Programming (group 1) | IP1 | Computer | 1 | 144 | Medium |
| Introduction to Programming (group 2) | IP2 | Computer | 1 | 145 | High |
| Programming Methodology (group 1) | PM1 | Computer | 1 | 114 | Medium |
| Programming Methodology (group 2) | PM2 | Computer | 1 | 119 | High |
| Professional Computer Tools | PCT | Computer | 1 | 124 | Medium |
| Databases | DB | Computer | 2 | 58 | Medium |
| Human Computer Interfaces | HCI | Computer | 2 | 260 | High |
| Information Systems | InS | Computer | 2 | 188 | Medium |
| Software Engineering | SE | Computer | 2 | 58 | Medium |
| Interactive Systems | IS | Computer | 3 | 84 | High |
| Requirement Engineering | RE | Computer | 3 | 36 | Medium |
| Software Design and Construction | SDC | Computer | 3 | 50 | Medium |
| Introduction to Computer Science | ICS1 | Electrical Engineering | 1 | 100 | Low |
| Introduction to Computer Science | ICS2 | Electronic Engineering | 1 | 198 | High |
| Introduction to Computer Science | ICS3 | Civil Engineering | 1 | 85 | Low |
| Introduction to Computer Science | ICS4 | Mining Engineering | 1 | 77 | Low |

We divided or grouped our 16 different courses (see Table 1) into three usage levels of Moodle activities in courses (see Table 2). Moodle provides us resources (text and web page, link to files or websites, and label) and different types of activities (assignments, chat, choice, database, forum, glossary, lesson, quiz, survey, wiki, workshop, etc.). We have defined three levels of usage by the number of activities used in the course:

· **Low level**: The course only has one or no activity.
· **Medium level**: The course has two different types of activities.

· **High level**: The course has three or more different types of activities.

Moodle provides a wide range of activities such as Assignments, Databases, Chats, Choice, Questionnaires, Quiz, Surveys, Forums, Glossaries, Lessons, SCORM packages, Workshops, Wikis, etc.). The most frequent activities in our courses are Assignments, Forums, and Quizzes. So, normally low-level courses only use one of these three activities, medium level two of them, and high level three or more activities. Table 2 shows the number of courses in each group grouped by usage level.

**Table 2.** List of groups by Moodle usage

| No. | Group  | No. of subjects |
|-----|--------|-----------------|
| 1   | High   | 5               |
| 2   | Medium | 8               |
| 3   | Low    | 3               |

We also propose our ontology for defining 5 high-level attributes starting from 58 low-level attributes or actions provided by Moodle logs (see Table 3).

**Table 3.** Ontology and Moodle low-level actions of each category.

| LEARNING / READING / VIEWING | COMMUNICATING | WORKING/DOING | EVALUATING / EXAMINING | ENGAGEMENT |
|---|---|---|---|---|
| blog view | forum add discussion | assignment upload | hotpot submit | Number of total interactions |
| book view all | forum add post | assignment view | hotpot view | |
| course enrol | forum search | assignment view all | hotpot view all | Number of Days connected |
| course recent | forum subscribe | assignment view submission | questionnaire submit | |
| course user report | forum subscribe all | choice choose | questionnaire update | |
| course view | forum update | choice choose again | questionnaire view | |
| folder view | forum update post | choice view | questionnaire view all | |
| folder view all | forum user report | choice view all | quiz attempt | |
| imscp view all | forum view discussion | teamwork update | quiz close attempt | |
| page view | forum view forum | teamwork view | quiz continue attempt | |
| page view all | forum view forums | teamwork view all | quiz continue attempt | |
| resource view | wiki edit | | quiz preview | |
| resource view all | wiki update | | quiz review | |
| url view | wiki view | | quiz view | |
| url view all | wiki view all | | quiz view all | |

As depicted in Table 3, our ontology generalizes the 58 raw/low-level events provided by the Moodle logs into only five attributes or high-level categories. The first category references all the actions about consulting resources (LEARNING/READING/VIEWING), the second groups the students' communication events (COMMUNICATING), the third deals with the students' work

(WORKING/DOING), the fourth is about students' evaluation (EXAMINING/EVALUATING) and the last is about the students' general ENGAGEMENT in the course. The first four attributes of our ontology are a number (from 0 to 100) that is the percentage of events of each type that each student has done in Moodle. The last attribute is the most general and is also a number (between 0 and 100) obtained from the total number of interactions/events and the number of days connected.

Finally, we have created two different datasets or data files: one with the original previously-described numerical data, and the other discretizing the attributes in two labels (HIGH and LOW) by using the equal width discretization method.

In both cases, we added a new attribute or class to predict at the end of our 5 attributes. This class is the final mark obtained by the students in the course, which is the value to predict in a classification task. The final mark (value between 0 and 10) has been discretized into two values or labels: FAIL if the student's final mark is lower than 5 or PASS if the students' final mark is higher than 5.

### 3.2. Methodology for experimentation

The methodology used in our experimentation consisted of these steps (see Figure 1):

- Firstly, we downloaded and preprocessed the Moodle log in order to obtain both the numerical and discretized datasets for each course. We used a specific Java tool that we developed for doing this specific transformation task (López-Zambrano et al., 2020).
- Secondly, we executed the well-known J48 classification algorithm provided by the WEKA data mining environment for each one of the previous numerical and categorical datasets of 16 subjects or different courses. In this step, we obtained one prediction model for each course.
- Then, we grouped our 16 subjects/courses into 3 groups depending on their level of usage of Moodle activities (see Table 1).
- Next, we repeated the next two actions. We selected each prediction model obtained in one course one by one and we applied it to testing the datasets of all the other courses in the same group. We repeated this process with all the models and with all the datasets for each group.
- Finally, we obtained the values of the two evaluation metrics that we used (the area under the ROC Curve and AUC loss) when applying the prediction model for one course/subject over the other datasets in the same group. And we compared the results obtained when using the original raw low-level

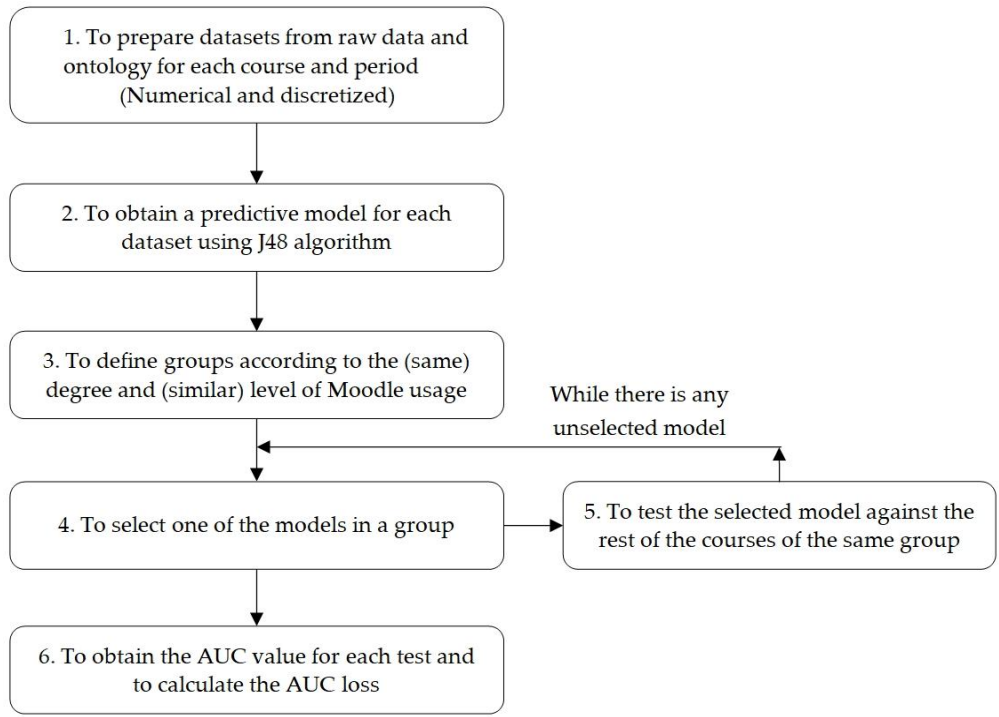

**Figure 1.** Methodology used in our experimentation.

## 4. Results

The results of these three groups are set out below (summarised in Table 3). Two experiments were conducted for each group, applying the J48 algorithm with balanced numerical and discretized datasets. These experiments consisted of having a first set of experiments for which high-level datasets were constructed (ontology) and a second experiment with datasets built with low-level attributes.

For each experiment (within the same group), we conducted an analysis of the best AUC obtained and the lowest error rate, or loss of portability, of the model. Thus, the results consist of two tables. At the top, a matrix is shown with the results of the AUC metric, obtained from the list of the general model for each subject (rows), compared to the average AUC for the individual datasets from each period for a subject (columns). The values of the main diagonal represent the testing of the general model for subjects over their own datasets, where this value is the reference AUC value (highest value), with regard to the AUCs from the other subjects. The second matrix (bottom) displays the difference between the highest AUC (reference) by row, with regard to each individual AUC. These values tell us how much precision is lost in the AUC when this model is tested with other subjects (portability), aiming to highlight the lowest values, as they indicate the lowest error rate or loss in the process of model portability or transferability.

### 4.1. Group of courses with high-level usage

For the high-level group, we can see in Table 4 that of the two tests, the best general results (averages) are in the datasets with ontology, revealing that the AUC average for numerical datasets is 0.62 and the average for discretized datasets is 0.61, higher than their equivalents in the tests without ontology. While there is only a small difference, the loss rate or difference in transferability does denote a greater difference, and within the same test group, the difference between numerical and discretized datasets is highly significant, where the tests with discretized data are much better.

If we focus on the tests with the best results, we can see that the best value for the AUC metric (0.675) is in the ICS2 subject obtained with discretized data and tested with the subject HCI. This is not concordant with the general average of the AUCs, whose highest value is for the numerical sets (0.62), with a tiny difference of one one-hundredth. However, it is concordant with the fact that the best rate of loss or difference is with the discretized data (0.10). We can also see that the generalized model obtained with the aforesaid subject (ICS2) has very good results, which is proven in the general averages (row) in both tests (with and without ontology).

With regard to the model that obtained the best average in the precision loss rate, we can see in Figure 2, the decision tree, defining the attribute COMMUNICATING (from the five ontology attributes – Table 1) as the attribute with the highest increase in information, which would define the prediction for a student passing the course.

```
J48 pruned tree
------------------

COMMUNICATING = LOW: Fail
COMMUNICATING = HIGH: Pass

Number of Leaves:       2
Size of the tree:       3
```

**Figure 2.** The best model for the high-level group with discretized dataset – Subject ICS2

**Table 4.** AUC results and loss of transferability (difference) with J48 – high-level group

**With Ontology**

| Course | AUC (Numerical Datasets) | | | | | | AUC (Discretized Datasets) | | | | | |
|---|---|---|---|---|---|---|---|---|---|---|---|---|
| | HCI | IS | ICS2 | IP2 | PM2 | avg | HCI | IS | ICS2 | IP2 | PM2 | avg |
| HCI | 0,890 | 0,511 | 0,592 | 0,535 | 0,528 | 0,61 | 0,710 | 0,672 | 0,608 | 0,614 | 0,624 | 0,65 |
| IS | 0,488 | 0,886 | 0,498 | 0,555 | 0,629 | 0,61 | 0,509 | 0,672 | 0,512 | 0,576 | 0,629 | 0,58 |
| ICS2 | 0,602 | 0,600 | 0,799 | 0,639 | 0,661 | 0,66 | 0,675 | 0,633 | 0,717 | 0,632 | 0,662 | 0,66 |
| IP2 | 0,483 | 0,484 | 0,589 | 0,849 | 0,550 | 0,59 | 0,536 | 0,651 | 0,512 | 0,704 | 0,630 | 0,61 |
| PM2 | 0,501 | 0,591 | 0,483 | 0,544 | 0,909 | 0,61 | 0,501 | 0,560 | 0,562 | 0,577 | 0,666 | 0,57 |
| | | | | | avg mean | 0,62 | | | | | avg mean | 0,61 |

| Course | AUC LOSS (Numerical Datasets) | | | | | | AUC LOSS (Discretized Datasets) | | | | | |
|---|---|---|---|---|---|---|---|---|---|---|---|---|
| | HCI | IS | ICS2 | IP2 | PM2 | avg | HCI | IS | ICS2 | IP2 | PM2 | avg |
| HCI | - | 0,379 | 0,298 | 0,355 | 0,362 | 0,35 | - | 0,038 | 0,102 | 0,095 | 0,085 | 0,08 |
| IS | 0,398 | - | 0,388 | 0,331 | 0,257 | 0,34 | 0,163 | - | 0,159 | 0,096 | 0,043 | 0,12 |
| ICS2 | 0,197 | 0,199 | - | 0,160 | 0,138 | 0,17 | 0,042 | 0,083 | - | 0,085 | 0,055 | 0,07 |
| IP2 | 0,366 | 0,364 | 0,260 | - | 0,298 | 0,32 | 0,169 | 0,053 | 0,192 | - | 0,074 | 0,12 |
| PM2 | 0,408 | 0,318 | 0,426 | 0,364 | - | 0,38 | 0,166 | 0,107 | 0,104 | 0,089 | - | 0,12 |
| | | | | | avg mean | 0,31 | | | | | avg mean | 0,10 |

**Without Ontology**

| Course | AUC (Numerical Datasets) | | | | | | AUC (Discretized Datasets) | | | | | |
|---|---|---|---|---|---|---|---|---|---|---|---|---|
| | HCI | IS | ICS2 | IP2 | PM2 | avg | HCI | IS | ICS2 | IP2 | PM2 | avg |
| HCI | 0,943 | 0,510 | 0,522 | 0,538 | 0,524 | 0,61 | 0,769 | 0,621 | 0,569 | 0,417 | 0,570 | 0,59 |
| IS | 0,485 | 0,927 | 0,494 | 0,470 | 0,606 | 0,60 | 0,479 | 0,816 | 0,577 | 0,555 | 0,656 | 0,62 |
| ICS2 | 0,514 | 0,590 | 0,783 | 0,500 | 0,569 | 0,59 | 0,503 | 0,558 | 0,619 | 0,485 | 0,516 | 0,54 |
| IP2 | 0,484 | 0,420 | 0,472 | 0,862 | 0,490 | 0,55 | 0,519 | 0,576 | 0,535 | 0,761 | 0,491 | 0,58 |
| PM2 | 0,514 | 0,489 | 0,530 | 0,618 | 0,899 | 0,61 | 0,574 | 0,488 | 0,522 | 0,592 | 0,793 | 0,59 |
| | | | | | avg mean | 0,59 | | | | | avg mean | 0,58 |

| Course | AUC LOSS (Numerical Datasets) | | | | | | AUC LOSS (Discretized Datasets) | | | | | |
|---|---|---|---|---|---|---|---|---|---|---|---|---|
| | HCI | IS | ICS2 | IP2 | PM2 | avg | HCI | IS | ICS2 | IP2 | PM2 | avg |
| HCI | - | 0,432 | 0,421 | 0,404 | 0,418 | 0,42 | - | 0,148 | 0,201 | 0,352 | 0,200 | 0,23 |
| IS | 0,442 | - | 0,433 | 0,457 | 0,321 | 0,41 | 0,337 | - | 0,238 | 0,260 | 0,160 | 0,25 |
| ICS2 | 0,270 | 0,193 | - | 0,283 | 0,215 | 0,24 | 0,116 | 0,061 | - | 0,134 | 0,103 | 0,10 |
| IP2 | 0,378 | 0,441 | 0,390 | - | 0,371 | 0,40 | 0,242 | 0,184 | 0,225 | - | 0,269 | 0,23 |
| PM2 | 0,385 | 0,410 | 0,369 | 0,281 | - | 0,36 | 0,219 | 0,305 | 0,271 | 0,200 | - | 0,25 |
| | | | | | avg mean | 0,37 | | | | | avg mean | 0,21 |

### 4.2. Group of courses with medium-level usage

For the medium-level group, we can see in Table 5 that of the two tests, the best general results (averages) are in the dataset tests with ontology, revealing that the AUC average for numerical datasets is 0.60 and the average for discretized datasets is 0.59, higher than their equivalents in the tests without ontology. There is a small difference, although the loss rate or difference in transferability does denote a greater difference, and within the same test group, the difference between numerical and discretized datasets is highly significant, where the tests with discretized data are much better.

If we focus on the tests with the best results, we see that the best value for the AUC metric (0.718) is for the subject SDC, obtained with numerical data and tested with the subject RE, which is concordant with the general AUC average, whose highest value is in the numerical tests (0.60), with is a small difference of one one-hundredth. However, it is not concordant with the fact that the best loss or difference rate is for discretized data (0.18). We also see that in the generalized model within the tests with discretized data, the subject PM1 has a good result in the general average for the loss rate (row) in the tests without ontology, although in the tests with ontology (employing a generalized model of high-level attributes), there are also good results in the IP1 and INS subjects, which share the same value of 0.13, six one-hundredths more, but still within the ideal value for good transferability of the model.

With regard to the subjects with the best average loss rate, Figure 3 shows that the decision tree defines the attributes LEARNING/READING/VIEWING and COMMUNICATING (from the five ontology results – Table 1) as the attributes with the greatest gain in information, defining that if there is a high level of LEARNING/READING/VIEWING, the student will pass or, conversely, if it is low, but with a high level of interaction in COMMUNICATING, the student will also pass.

Concerning the decision tree shown in Figure 4, it defines the attributes LEARNING/READING/VIEWING, COMMUNICATING, WORKING/DOING and EVALUATING/EXAMINING (from the five ontology attributes – Table 1) as the attributes with the greatest increase in information, once again defining that if there is a high level of LEARNING/READING/VIEWING, the student will pass or, conversely, if it is low, but with a high level of interaction in COMMUNICATING, the student will also pass. If the COMMUNICATING level is low, but the level of WORKING/DOING is high, then the student would pass, but if it is not high, then the student will only pass if the EVALUATING/EXAMINING level is high.

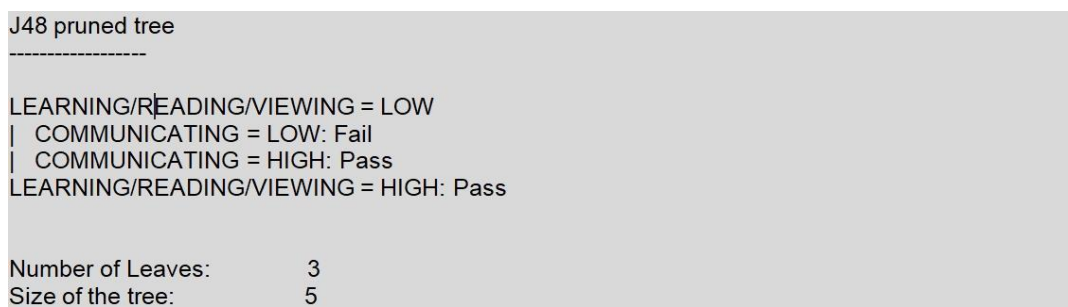

**Figure 3.** Best model for the medium-level group with discretized dataset – Subject IP1

```
J48 pruned tree
------------------

LEARNING/READING/VIEWING = LOW
|   COMMUNICATING = LOW
|   |   WORKING/DOING = LOW
|   |   |   EVALUATING/EXAMINING = LOW: Fail
|   |   |   EVALUATING/EXAMINING = HIGH: Pass
|   |   WORKING/DOING = HIGH: Pass
|   COMMUNICATING = HIGH: Pass
LEARNING/READING/VIEWING = HIGH: Pass

Number of Leaves:       5
Size of the tree:       9
```

**Figure 4.** Best model for the medium-level group with discretized dataset – Subject InS

**Table 5.** AUC results and loss of transferability (difference) with J48 – Medium-level group

### With Ontology

| Course | AUC (Numerical Datasets) | | | | | | | | | AUC (Discretized Datasets) | | | | | | | | |
|---|---|---|---|---|---|---|---|---|---|---|---|---|---|---|---|---|---|---|
| | IP1 | PM1 | DB | SDC | PCT | RE | SE | InS | avg | IP1 | PM1 | DB | SDC | PCT | RE | SE | InS | avg |
| IP1 | 0,835 | 0,567 | 0,589 | 0,552 | 0,508 | 0,589 | 0,620 | 0,582 | 0,61 | 0,772 | 0,637 | 0,621 | 0,601 | 0,688 | 0,643 | 0,643 | 0,652 | 0,66 |
| PM1 | 0,519 | 0,821 | 0,540 | 0,520 | 0,530 | 0,510 | 0,550 | 0,567 | 0,57 | 0,634 | 0,763 | 0,532 | 0,604 | 0,562 | 0,510 | 0,521 | 0,602 | 0,59 |
| DB | 0,670 | 0,623 | 0,980 | 0,590 | 0,571 | 0,566 | 0,521 | 0,640 | 0,65 | 0,612 | 0,583 | 0,775 | 0,555 | 0,616 | 0,567 | 0,543 | 0,551 | 0,60 |
| SDC | 0,502 | 0,596 | 0,516 | 0,788 | 0,469 | 0,718 | 0,549 | 0,504 | 0,58 | 0,474 | 0,562 | 0,628 | 0,696 | 0,505 | 0,590 | 0,480 | 0,551 | 0,56 |
| PCT | 0,633 | 0,621 | 0,611 | 0,610 | 0,911 | 0,641 | 0,572 | 0,670 | 0,66 | 0,592 | 0,564 | 0,577 | 0,582 | 0,812 | 0,567 | 0,581 | 0,582 | 0,61 |
| RE | 0,494 | 0,519 | 0,497 | 0,643 | 0,476 | 0,869 | 0,527 | 0,512 | 0,57 | 0,589 | 0,591 | 0,520 | 0,583 | 0,572 | 0,801 | 0,563 | 0,571 | 0,60 |
| SE | 0,540 | 0,510 | 0,520 | 0,560 | 0,530 | 0,511 | 0,962 | 0,523 | 0,58 | 0,527 | 0,562 | 0,550 | 0,588 | 0,504 | 0,614 | 0,694 | 0,548 | 0,57 |
| InS | 0,608 | 0,580 | 0,591 | 0,563 | 0,508 | 0,560 | 0,562 | 0,815 | 0,60 | 0,648 | 0,635 | 0,549 | 0,640 | 0,471 | 0,369 | 0,529 | 0,677 | 0,56 |
| | | | | | | | | avg mean | 0,60 | | | | | | | | avg mean | 0,59 |

| Course | AUC LOSS (Numerical Datasets) | | | | | | | | | AUC LOSS (Discretized Datasets) | | | | | | | | |
|---|---|---|---|---|---|---|---|---|---|---|---|---|---|---|---|---|---|---|
| | IP1 | PM1 | DB | SDC | PCT | RE | SE | InS | avg | IP1 | PM1 | DB | SDC | PCT | RE | SE | InS | avg |
| IP1 | - | 0,267 | 0,246 | 0,283 | 0,327 | 0,246 | 0,215 | 0,253 | 0,26 | - | 0,135 | 0,151 | 0,172 | 0,084 | 0,129 | 0,129 | 0,120 | **0,13** |
| PM1 | 0,302 | - | 0,281 | 0,301 | 0,291 | 0,311 | 0,271 | 0,254 | 0,29 | 0,129 | - | 0,231 | 0,159 | 0,201 | 0,253 | 0,242 | 0,162 | 0,20 |
| DB | 0,310 | 0,357 | - | 0,390 | 0,409 | 0,414 | 0,459 | 0,340 | 0,38 | 0,163 | 0,192 | - | 0,220 | 0,159 | 0,208 | 0,232 | 0,224 | 0,20 |
| SDC | 0,286 | 0,192 | 0,272 | - | 0,319 | 0,070 | 0,239 | 0,284 | **0,24** | 0,222 | 0,134 | 0,068 | - | 0,191 | 0,107 | 0,216 | 0,145 | 0,15 |
| PCT | 0,278 | 0,290 | 0,300 | 0,301 | - | 0,270 | 0,339 | 0,241 | 0,29 | 0,220 | 0,248 | 0,235 | 0,230 | - | 0,245 | 0,231 | 0,230 | 0,23 |
| RE | 0,375 | 0,350 | 0,372 | 0,226 | 0,393 | - | 0,342 | 0,357 | 0,34 | 0,212 | 0,210 | 0,281 | 0,218 | 0,229 | - | 0,238 | 0,230 | 0,23 |
| SE | 0,422 | 0,452 | 0,442 | 0,402 | 0,432 | 0,451 | - | 0,439 | 0,43 | 0,167 | 0,132 | 0,144 | 0,107 | 0,190 | 0,080 | - | 0,146 | 0,14 |
| InS | 0,207 | 0,235 | 0,224 | 0,252 | 0,307 | 0,255 | 0,253 | - | 0,25 | 0,029 | 0,042 | 0,128 | 0,038 | 0,206 | 0,309 | 0,148 | - | **0,13** |
| | | | | | | | | avg mean | 0,31 | | | | | | | | avg mean | 0,18 |

### Without Ontology

| Course | AUC (Numerical Datasets) | | | | | | | | | AUC (Discretized Datasets) | | | | | | | | |
|---|---|---|---|---|---|---|---|---|---|---|---|---|---|---|---|---|---|---|
| | IP1 | PM1 | DB | SDC | PCT | RE | SE | InS | avg | IP1 | PM1 | DB | SDC | PCT | RE | SE | InS | avg |
| IP1 | 0,938 | 0,588 | 0,542 | 0,545 | 0,610 | 0,493 | 0,579 | 0,523 | 0,60 | 0,811 | 0,441 | 0,496 | 0,535 | 0,500 | 0,500 | 0,414 | 0,510 | 0,53 |
| PM1 | 0,496 | 0,689 | 0,589 | 0,478 | 0,567 | 0,624 | 0,484 | 0,486 | 0,55 | 0,476 | 0,585 | 0,458 | 0,550 | 0,515 | 0,564 | 0,512 | 0,559 | 0,53 |
| DB | 0,495 | 0,491 | 0,976 | 0,535 | 0,457 | 0,670 | 0,581 | 0,517 | 0,59 | 0,551 | 0,500 | 0,652 | 0,551 | 0,476 | 0,500 | 0,510 | 0,499 | 0,53 |
| SDC | 0,492 | 0,518 | 0,467 | 0,809 | 0,504 | 0,558 | 0,496 | 0,456 | 0,54 | 0,532 | 0,593 | 0,430 | 0,924 | 0,531 | 0,610 | 0,484 | 0,622 | 0,59 |
| PCT | 0,459 | 0,496 | 0,337 | 0,585 | 0,891 | 0,612 | 0,382 | 0,492 | 0,53 | 0,494 | 0,500 | 0,447 | 0,567 | 0,712 | 0,553 | 0,470 | 0,551 | 0,54 |
| RE | 0,439 | 0,524 | 0,329 | 0,553 | 0,579 | 0,956 | 0,473 | 0,577 | 0,55 | 0,568 | 0,543 | 0,529 | 0,614 | 0,508 | 0,756 | 0,545 | 0,569 | 0,58 |
| SE | 0,526 | 0,581 | 0,611 | 0,559 | 0,486 | 0,614 | 0,964 | 0,494 | 0,60 | 0,487 | 0,500 | 0,500 | 0,375 | 0,473 | 0,431 | 0,718 | 0,451 | 0,49 |
| InS | 0,484 | 0,495 | 0,671 | 0,583 | 0,486 | 0,610 | 0,533 | 0,704 | 0,57 | 0,526 | 0,500 | 0,429 | 0,625 | 0,528 | 0,454 | 0,500 | 0,761 | 0,54 |
| | | | | | | | | avg mean | 0,57 | | | | | | | | avg mean | 0,54 |

| Course | AUC LOSS (Numerical Datasets) | | | | | | | | | AUC LOSS (Discretized Datasets) | | | | | | | | |
|---|---|---|---|---|---|---|---|---|---|---|---|---|---|---|---|---|---|---|
| | IP1 | PM1 | DB | SDC | PCT | RE | SE | InS | avg | IP1 | PM1 | DB | SDC | PCT | RE | SE | InS | avg |
| IP1 | - | 0,350 | 0,396 | 0,393 | 0,328 | 0,446 | 0,359 | 0,415 | 0,38 | - | 0,370 | 0,315 | 0,277 | 0,311 | 0,311 | 0,397 | 0,301 | 0,33 |
| PM1 | 0,193 | - | 0,100 | 0,211 | 0,122 | 0,065 | 0,205 | 0,203 | 0,16 | 0,108 | - | 0,127 | 0,035 | 0,070 | 0,021 | 0,073 | 0,025 | 0,07 |
| DB | 0,481 | 0,485 | - | 0,441 | 0,519 | 0,307 | 0,395 | 0,459 | 0,44 | 0,101 | 0,152 | - | 0,101 | 0,176 | 0,152 | 0,142 | 0,153 | 0,14 |
| SDC | 0,317 | 0,291 | 0,342 | - | 0,305 | 0,252 | 0,313 | 0,353 | 0,31 | 0,392 | 0,331 | 0,494 | - | 0,393 | 0,314 | 0,440 | 0,302 | 0,38 |
| PCT | 0,432 | 0,395 | 0,554 | 0,306 | - | 0,279 | 0,509 | 0,399 | 0,41 | 0,218 | 0,212 | 0,265 | 0,145 | - | 0,159 | 0,242 | 0,161 | 0,20 |
| RE | 0,517 | 0,432 | 0,627 | 0,403 | 0,377 | - | 0,483 | 0,379 | 0,46 | 0,188 | 0,213 | 0,227 | 0,142 | 0,248 | - | 0,211 | 0,187 | 0,20 |
| SE | 0,438 | 0,383 | 0,353 | 0,405 | 0,478 | 0,351 | - | 0,470 | 0,41 | 0,231 | 0,218 | 0,218 | 0,343 | 0,245 | 0,287 | - | 0,267 | 0,26 |
| InS | 0,221 | 0,209 | 0,033 | 0,121 | 0,218 | 0,094 | 0,171 | - | **0,15** | 0,235 | 0,261 | 0,332 | 0,136 | 0,233 | 0,307 | 0,261 | - | 0,25 |
| | | | | | | | | avg mean | 0,34 | | | | | | | | avg mean | 0,23 |

### 4.3. Group of courses with low-level usage

For the low-level group, we can see in Table 6 that of the two tests, the best general results (averages) are in the dataset tests with ontology, revealing that the AUC averages for numerical datasets is 0.63 and the average for discretized datasets is 0.61, higher than their equivalents in the tests without ontology. There is a small difference, although the loss rate or difference in transferability does denote a greater difference, and within the same test group, the difference between numerical and discretized datasets is highly significant, where the tests with discretized data are much better.

If we now focus on the tests with the best results, we can see that the best value for the AUC metric (0.683) is in the ICS1 subject obtained with discretized data and tested with the subject ICS4. This is not concordant with the general average of the AUCs whose highest value is for the numerical sets (0.63), with a small difference of two one-hundredths. However, it is concordant with the fact that the best rate of loss or difference is with the discretized data (0.13). We can also observe that the generalized model obtained with the aforesaid subject (ICS1) has very good results, which is proven in the general averages (row) for the matrix of discretized data with ontology. The value is 0.07, which is below the ideal for determining a good transfer of the model, in this case for a general model with high-level attributes.

With regard to the subject with the best average loss rate, we can see in Figure 5, the decision tree, defining the attribute COMMUNICATING (from the five attributes of ontology – Table 1) as the attribute with the highest increase in information, which would define the predictability for a student passing the course.

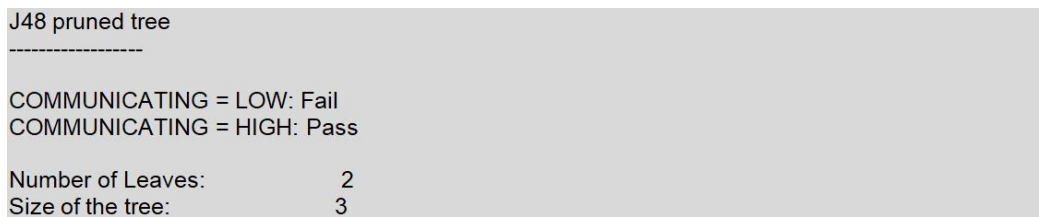

**Figure 5.** Best model for the low-level group with discretized dataset – Subject ICS1

**Table 6.** AUC results and loss of transferability (difference) with J48 – Low-level group

**With Ontology**

| Course | AUC (Numerical Datasets) | | | | Discretized Datasets | | | |
|---|---|---|---|---|---|---|---|---|
| | ICS1 | ICS3 | ICS4 | avg | ICS1 | ICS3 | ICS4 | avg |
| ICS1 | 0,860 | 0,592 | 0,500 | 0,65 | 0,722 | 0,615 | 0,683 | 0,67 |
| ICS3 | 0,506 | 0,820 | 0,560 | 0,63 | 0,512 | 0,750 | 0,565 | 0,61 |
| ICS4 | 0,510 | 0,531 | 0,832 | 0,62 | 0,500 | 0,500 | 0,600 | 0,53 |
| | | | avg mean | 0,63 | | | avg mean | 0,61 |

| Course | AUC LOSS (Numerical Datasets) | | | | AUC LOSS (Discretized Datasets) | | | |
|---|---|---|---|---|---|---|---|---|
| | ICS1 | ICS3 | ICS4 | avg | ICS1 | ICS3 | ICS4 | avg |
| ICS1 | - | 0,268 | 0,360 | 0,31 | - | 0,107 | 0,039 | 0,07 |
| ICS3 | 0,314 | - | 0,260 | 0,29 | 0,239 | - | 0,186 | 0,21 |
| ICS4 | 0,322 | 0,301 | - | 0,31 | 0,100 | 0,100 | - | 0,10 |
| | | | avg mean | 0,30 | | | avg mean | 0,13 |

**Without Ontology**

| Course | AUC (Numerical Datasets) | | | | Discretized Datasets | | | |
|---|---|---|---|---|---|---|---|---|
| | ICS1 | ICS3 | ICS4 | avg | ICS1 | ICS3 | ICS4 | avg |
| ICS1 | 0,917 | 0,491 | 0,404 | 0,60 | 0,761 | 0,470 | 0,591 | 0,61 |
| ICS3 | 0,554 | 0,938 | 0,527 | 0,67 | 0,375 | 0,707 | 0,502 | 0,53 |
| ICS4 | 0,414 | 0,495 | 0,771 | 0,56 | 0,410 | 0,460 | 0,682 | 0,52 |
| | | | avg mean | 0,61 | | | avg mean | 0,55 |

| Course | AUC LOSS (Numerical Datasets) | | | | AUC LOSS (Discretized Datasets) | | | |
|---|---|---|---|---|---|---|---|---|
| | ICS1 | ICS3 | ICS4 | avg | ICS1 | ICS3 | ICS4 | avg |
| ICS1 | - | 0,426 | 0,513 | 0,47 | - | 0,291 | 0,170 | 0,23 |
| ICS3 | 0,384 | - | 0,411 | 0,40 | 0,333 | - | 0,205 | 0,27 |
| ICS4 | 0,357 | 0,277 | - | 0,32 | 0,273 | 0,222 | - | 0,25 |
| | | | avg mean | 0,39 | | | avg mean | 0,25 |

## 5. Conclusions

This paper aims to improve the portability or transferability of predictive models of students' performance by using an ontology that uses a taxonomy of actions on students' interactions with the Moodle learning management system. We compare the results of this new proposed approach against our previous results when we used low-level raw attributes directly obtained from Moodle logs. The results obtained show that the use of the proposed ontology significantly improves the portability of the models in terms of their predictive accuracy. So, the answer to our initial research question is yes, the ontological models obtained in one source course can be applied to other different target courses with similar usage levels without losing prediction accuracy.

One of the limitations of this work is the specific attributes/variables used in our proposed ontology. For example, it is also important to discuss if the "number of total interactions" are truly showing engagement when learning using LMS. The number of actions includes the behavior of supposed relevant activity in the LMS and were are assuming that all of these actions could indicate that the student is properly involved in his learning process. As traditionally happens with study time, however, this variable by itself is very tricky. It may seem that the more time those students spend studying, the better grades they should receive, but it is not that simple; it mainly depends on the quality of the study time, and something similar could be occurring with the relevant actions; more activity in the LMS does not assure better results (Cerezo et al., 2016).

Regarding the application of the results obtained in this work and the potential for using them within other domains; it is important to notice that currently there is an increasing interest in the generalization and portability of prediction models and specifically with Moodle LMS (Moollao-Olive et al., 2019). In this line, our proposal can be applied not only to Learning Management Systems as Moodle but also to other different domains or data sources such as Intelligent Tutoring Systems (ITSs), Massive Online Open Courses (MOOCs), Traditional face-to-face educational environments, Blended Learning and Multimodal Learning environments, and so on.

Finally, as a future study, we are currently working on:
- Using a higher number of courses with much more data/students from different areas/domains, not only engineering and computer science, but also fields such as science, biology, medicine, philosophy, and literature, in order to generalize the good results that we obtained in this study.
- Discovering predictive models that can be portable/transferable as soon as possible in the early stages of the course. This means we would not have to wait until the end of the course to have all Moodle usage data available, and the obtained models could be used as general early warning prediction models for different similar courses (Romero & Ventura, 2019).

**Acknowledgments**

This work would not be possible without the funding from the Ministry of Sciences and Innovation I+D+I TIN2017-83445-P.**Acknowledgments**

This work would not be possible without the funding from the Ministry of Sciences and Innovation I+D+I TIN2017-83445-P.